\documentclass[10pt,letterpaper]{article}
\usepackage[OT1]{fontenc}
\usepackage{times}
\usepackage[letterpaper,textwidth=5.5in,textheight=9in,top=1in]{geometry}
\usepackage{amsmath,amssymb}
\usepackage[authoryear,round]{natbib}

\usepackage{amsthm}
\newtheorem{proposition}{Proposition}
\newtheorem{lemma}{Lemma}

\theoremstyle{definition}
\newtheorem{definition}{Definition}
\usepackage{algorithm}
\usepackage{algpseudocode}
\usepackage{hyperref}
\usepackage{url}
\usepackage{graphicx}
\usepackage{booktabs,longtable,array,seqsplit,tabularx}
\usepackage{adjustbox}
\usepackage{microtype}
\usepackage{caption}
\usepackage{tikz}
\usepackage{orcidlink}
\usepackage{placeins}
\usepackage{float}
\usetikzlibrary{arrows.meta,positioning,calc}
\definecolor{auditblue}{HTML}{0072B2}

\newcommand{\paperci}[2]{\shortstack[c]{$#1$\\{\scriptsize$#2$}}}

\hypersetup{
  pdfauthor={Satwik Sai Prakash Sahoo, Bibhor Puhan, Shubhankar Mahanta},
  pdftitle={The Margin Knows the Background: Uneven Abstention in Frozen Vision-Language Models},
  hidelinks
}

\title{The Margin Knows the Background: Uneven Abstention in Frozen Vision-Language Models}

\author{%
  \normalsize Satwik Sai Prakash Sahoo\textsuperscript{1}\,\orcidlink{0009-0002-1943-9323}\quad
  Bibhor Puhan\textsuperscript{2}\,\orcidlink{0009-0007-0562-7258}\quad
  Shubhankar Mahanta\textsuperscript{2}\,\orcidlink{0009-0006-7314-6627}\\[6pt]
  \small\textsuperscript{1}Odisha University of Technology and Research\\
  \small\textsuperscript{2}Siksha `O' Anusandhan
}
\date{}

\begin{document}
\raggedbottom

\maketitle

\begin{abstract}
Selective prediction allows a model to abstain on uncertain inputs, but a single confidence threshold can disproportionately reject examples from particular groups. We study how confidence scores produce this imbalance in frozen vision-language models (OpenAI CLIP, LAION CLIP, SigLIP). We express a family of scores, including the margin and maximum logit, through the average and the difference of the two highest class logits, and ask which carries group information. We show that equal group coverage at every threshold is equivalent to independence between the score and group membership. A randomized margin baseline tests whether a reduction in coverage disparity beats what adding noise achieves. With score choices fixed before testing, the effect of choice depends on the dataset. On Waterbirds, the margin creates coverage gaps 2.5 to 5.5 times larger than maximum logit at 80\% coverage, its group signal is in the difference component, and paired background swaps show the background moves it, although maximum logit increases overall error. On NICO++, maximum logit instead increases worst-context error by 5.4 to 6.6 points and offers no improvement over the randomized baseline for OpenAI CLIP. Across both datasets, how far a score pushes a group down its ranking tracks the coverage gap (Spearman 0.88, 0.92) better than plain group identifiability (0.61, 0.65). These results explain why a score that abstains more evenly on one dataset may be less effective on another and motivate judging group coverage and selective risk together against randomized baselines.
\end{abstract}

\section{Introduction}
\label{sec:intro}
\begin{figure}[!htbp]
\centering
\includegraphics[width=\linewidth]{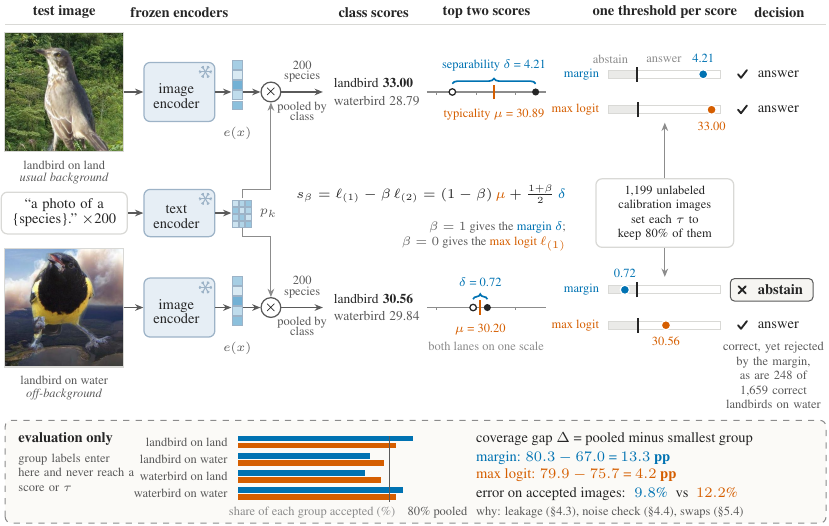}
\caption{The audit on two Waterbirds test images (OpenAI CLIP, protocol A). Each is the median-margin image of its case.}
\label{fig:architecture}
\end{figure}
Frozen vision-language models classify images by comparing them with text descriptions of candidate classes \citep{radford2021clip,zhai2023siglip}. A confidence threshold lets the classifier abstain, but its effects can differ across groups. Low average error among accepted inputs can coexist with poor coverage of a difficult group, and abstention can increase errors within some groups \citep{jones2021selective}. We ask which information in a frozen model's confidence score determines uneven acceptance.

The margin between the two largest class logits is a natural confidence score. Yet a large margin can reflect agreement between an object and its background, as well as evidence about the object itself. We separate the top-two logits into their average and their difference, called typicality and separability. The margin retains only separability; maximum logit combines both. This decomposition lets us compare scores without changing the encoder, prompts, or predicted label (Figure~\ref{fig:architecture}).

We relate group coverage to the score distribution and express a classical AUROC identity in coverage coordinates, distinguishing downward group shifts from group identifiability. We then audit three frozen encoders on Waterbirds and NICO++ using paired intervals and controlled background changes. A randomized margin baseline tests whether coverage gains exceed what noise achieves. Maximum logit reduces Waterbirds' coverage gaps, but its error cost and its comparison with randomness change on NICO++. The evidence supports a mechanism, not a universally preferred score.

\section{Related Work}
\label{sec:related}
\paragraph{Selective prediction and groups.}
Selective classification rejects predictions according to a confidence score \citep{geifman2017selective}. \citet{jones2021selective} show that it can magnify group disparities, while \citet{narasimhan2024reject} derive rejectors for balanced and worst-group error in long-tailed classification. Our focus is the coverage induced by alternative scores from a fixed VLM under one group-blind threshold. We report both coverage and risk, including whole-curve evaluation following \citet{traub2024evaluation}.

\paragraph{Backgrounds and frozen VLMs.}
Background dependence is established in image recognition \citep{xiao2021backgrounds} and zero-shot CLIP \citep{wang2024sober}. PerceptionCLIP marginalizes over inferred context \citep{an2024perceptionclip}, SAGE selects prompts using class separation \citep{ye2026sage}, and TIE translates image embeddings using estimated spurious directions \citep{lu2025tie}. These methods change classification. We retain the original predictions and study which inputs a confidence score accepts. Our background intervention uses a same-type control, following the motivation of \citet{xiao2021backgrounds}.

\paragraph{Confidence, prompts, and calibration.}
Prompt and image consistency can identify errors in multimodal classification \citep{ge2023improving} and selective visual question answering \citep{khan2024consistency}; PARC measures reliability under prompt variations \citep{schmalfuss2025parc}, and cross-prompt score variance flags fragile predictions in zero-shot VLM safety classification \citep{weng2026promptvariance}. Temperature scaling can improve VLM calibration under distribution shift \citep{tu2024vlmcalibration}, while memory-augmented selective prediction uses retrieved examples to stabilize representations and calibrate scores \citep{sarkar2026memory}. Recent work also documents class-conditional failures of conformal coverage under shift \citep{sharma2026marginal}. Conformal coverage concerns prediction-set inclusion, whereas our coverage is the fraction of inputs answered. We complement these studies by isolating the top-two score components, measuring group under-coverage and checking improvements against noise.

\section{Setup and Problem Formulation}
\label{sec:setup}

\subsection{Models}
\label{sec:models}
In our study, we have employed three VLM encoders: OpenAI CLIP ViT-B/32 \citep{radford2021clip}, a ViT-B/32 trained on LAION-2B \citep{schuhmann2022laion5b} and SigLIP ViT-B/16 \citep{zhai2023siglip}. All three are trained on different datasets, but what makes SigLIP different is that it is trained with a sigmoid loss per image-caption pair. So they are three different pretraining recipes, not three different random seeds of one model. All three models are frozen and are imported through the OpenCLIP library \citep{cherti2023openclip}. Each image is encoded one time, and every score in the paper is computed from that embedding and a fixed set of prompts.

Given an image encoder that outputs a vector $e(x)$ and a text encoder that outputs a vector $p_k$ of the same length for class $k$, we compute the class logit as $\ell_k(x) = \gamma\, e(x)^\top p_k + b$, and the prediction as $\hat{y} = \arg\max_k \ell_k(x)$. Every embedding is normalized to a length of 1, so the dot product is just the cosine similarity. Both CLIP models have the same scale and bias, $\gamma = 100$ and $b = 0$, whereas SigLIP has $\gamma = 117.3$ and $b = -12.9$, because SigLIP learns a bias term \citep{zhai2023siglip} and CLIP has none. We also do not pass the SigLIP logits to softmax, because it's not trained on softmax but rather on sigmoid. 

For our set of text prompts, we put the class name after the prefix ``a photo of a'' (for example, ``a photo of a waterbird''). For the Waterbirds dataset, we follow two protocols, A and B. A uses the 200 real bird species (for example, ``a photo of a Black footed Albatross.''),  computes 200 logits and with logsumexp $\ell_c(x) = \log \sum_{s \in c} \exp \ell_s(x)$, we pool the species logits into two classes; as a convention, the same pooling is also applied to SigLIP's species logits as well. In contrast, B simply uses ``waterbird'' and ``landbird'' as the suffix, and A acts as the main protocol as it is a stronger classifier. For the NICO++ dataset, there is only a single protocol, the 60 category names (for example, ``a photo of a car.''). 

\subsection{Data and groups}
\label{sec:data}
\begin{figure}[!htbp]
\centering
\includegraphics[width=\linewidth]{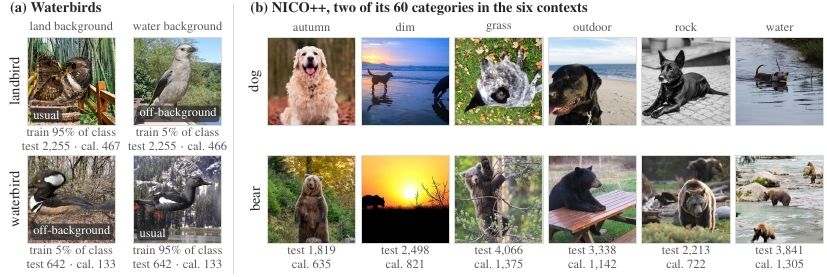}
\caption{Groups in Waterbirds (a) and NICO++ (b). Pictures are each group's median-margin test image (a) and the first test image per category and context without people or text (b).}
\label{fig:data}
\end{figure}
Waterbirds \citep{sagawa2020groupdro} pairs bird images with land or water backgrounds. We use 1,199 validation images for calibration and 5,794 test images for evaluation. Groups cross bird class with background, $g=2y+\mathrm{place}$ (Figure~\ref{fig:data}). We call mismatched groups ``off-background''; their minority status refers to the training distribution.

NICO++ \citep{zhang2023nico} supplies 60 categories across six common contexts, which define the groups. We use 17,775 test-list images and a fixed 6,000-image training-list sample for calibration. This audits shared contexts rather than leave-one-domain-out training. Table~\ref{tab:groupcounts} reports group sizes.

\subsection{Selective prediction protocol}
\label{sec:protocol}
Given a score $s(x)$, a selective classifier pairs the prediction of a model with that score and allows the model to answer only when $s(x) \ge \tau$ \citep{geifman2017selective} or else it abstains. Each image also has a group label $G$. Coverage represents the fraction of images that get answered, $C(\tau) = \Pr[s(X) \ge \tau]$. Selective risk measures the error rate among the non-abstained inputs, $R(\tau) = \Pr[\hat{y} \ne y \mid s(X) \ge \tau]$. Computing the same two within one group $g$ gives the group coverage $C_g(\tau)$ and group risk $R_g(\tau)$. Note that the model does not see $G$, and we use it only for evaluation purposes. 

A single threshold is used, fitted on unlabeled data. For Waterbirds, we use a validation split and for NICO++ we use 6,000 random fixed samples from the official training lists. We call this the calibration split. For a given target coverage $c$, we find the largest possible threshold $\tau_c$, the $\lceil cM \rceil$-th largest of the $M$ calibration scores, so that at least a fraction $c$ of the calibration split is answered. This same $\tau_c$ is then applied to the test split and to every group. Our selected grid is $c \in \{0.7, 0.8, 0.9\}$ with 0.8 as the main coverage.

At each $c$ we report the worst-group risk $\max_g R_g(\tau_c)$, and we compute the coverage gap by subtracting the smallest group coverage from the pooled coverage:
\begin{equation}
\Delta(\tau) = C(\tau) - \min_g C_g(\tau),
\label{eq:gap}
\end{equation}

where $\Delta(\tau)\ge\; 0$, and it measures the gap between the least answered group and the pool. In some cases, the worst group's error rate is low because the model abstained on most of that group, and hence the coverage gap helps in identifying those cases. We only report the group-risk if at least 25 of the group-images are answered; we mark it as ``missing'' if it is below 25. Including this, we also give each group risk a Wilson 95\% interval \citep{brown2001binomial}, to show the range where the group's true error rate plausibly lies. Following \citet{jones2021selective}, we also compare each group's risk with a group-agnostic reference, meaning a fair ``rule'' where this rule only notes the model's correct and wrong predictions without seeing their group, and answers as many images with the same overall error rate; then we evaluate if the real score is doing poorly in that group by checking whether the error rate is greater than this rule's error rate or not. We also compute the AUGRC \citep{traub2024evaluation} for the whole risk-coverage curve.

\subsection{Protocol freeze and uncertainty}
\label{sec:freeze}
We use a test lock for the final evaluation, while disclosing the earlier pipeline-validation read in the footnote below. All development decisions were made on the validation split. We ran Waterbirds first and fixed the scene vocabulary, the templates, the weights, coverage grid, scores, and the 25-image rule on the Waterbirds validation split and then read the Waterbirds test split once.\footnote{The test split had been run once before the freeze, with an earlier vocabulary, while the pipeline was being validated. No setting was changed on the basis of that run.} NICO++ was opened only after this freeze, with every setting carried over unchanged, and was run once. The subsequent diagnostics of Section~\ref{sec:results} changed no deployed score or threshold.

We have also implemented a bootstrap with a stratified approach where we retain the same group size while resampling within each group. Resamples in which a group falls under 25 accepted images are counted and left out of the worst-group risk percentiles. We resample $B = 2000$ times and take the 2.5th and 97.5th percentiles of the $B$ values as a 95\% interval, we keep the threshold at their calibration values, because under a chosen threshold, we want the interval to reflect only the finite size of the test split. When a score is compared to the margin, both are computed on the same resample, and the interval is taken on their difference. We also resampled the calibration split and refitted every threshold on each resample, and it turns out that it widens the interval by about a tenth. The bootstrap considers resampling source birds instead of composites for background swap (Section~\ref{sec:swap}), because the  four composites of one bird share its foreground and are not independent.

\section{Methodology}
\label{sec:scores}

\subsection{Confidence scores and their decomposition}
\label{sec:score-menu}
We compare several confidence scores, for example, the max logit or the winner class's logit \citep{hendrycks2022scaling}, the margin which measures the gap between the winner and the follow-up logit, the negative energy \citep{liu2020energy} a smooth version of the max logit that sums over all classes, the maximum softmax (MSP) \citep{hendrycks2017msp} which is the largest probability, a prompt-probe baseline in the SAGE separation \citep{ye2026sage}, the template variance, the template consistency of \citet{ge2023improving}, image-probe scores, which apply four augmentations per image, hence four extra encoder passes each \citep{bahat2018confidence}. Finally, we include our context-stability score, which did not help. On the two classes of Waterbirds, max softmax and entropy rank the image exactly as the margin does. We take the winner logit $\ell_{(1)}$, and runner-up logit $\ell_{(2)}$ to express separability $\delta$ and typicality $\mu$ as follows:
\begin{equation}
\mu = \tfrac12\big(\ell_{(1)} + \ell_{(2)}\big),\quad \delta = \ell_{(1)} - \ell_{(2)},
\label{eq:components}
\end{equation}
 where $\ell_{(1)} \ge \ell_{(2)}$ by definition. Here, $\mu$ is the average of the top two logits, how strongly the picture matches the prompts at all, and $\delta$'s definition is the same as the margin. By combining these two equations we get the following:
\begin{equation}
s_\beta = \ell_{(1)} - \beta\,\ell_{(2)} = (1-\beta)\,\mu + \tfrac{1+\beta}{2}\,\delta
\label{eq:family}
\end{equation}
Here $\beta$ works as a seesaw on how much the score should depend on separability and typicality. Here $\beta = 1$ means all margin,  $\beta = 0$ is the max logit, and $\beta = -1$ means all typicality or the logits sum. 
\subsection{Coverage and group dependence}
\label{sec:proposition}
\begin{figure}[!htbp]
\centering
\includegraphics[width=\linewidth]{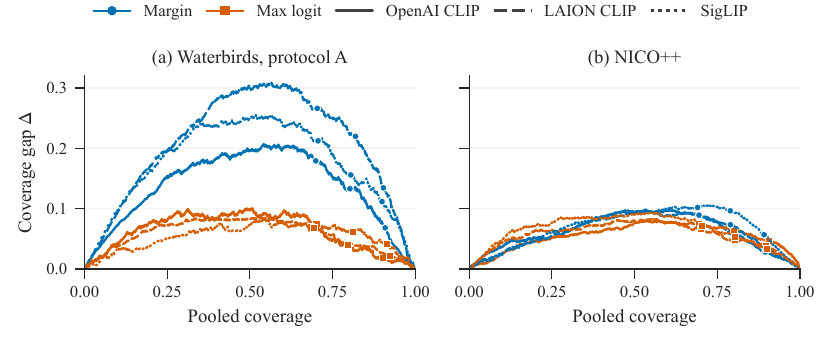}
\caption{Test-set coverage gap $\Delta$ against pooled coverage for the margin (blue circles) and maximum logit (orange squares). Line styles indicate OpenAI CLIP (solid), LAION CLIP (dashed), and SigLIP (dotted). The plots depict Waterbirds under protocol A (left) and NICO++ (right). Curves sweep over thresholds of test scores, with markers indicating thresholds tuned for $c\in\{0.7,0.8,0.9\}$. By Lemma~\ref{lem:area}, each curve has an area at least as large as the corresponding one-sided weighted leakage. The margin creates a larger gap across Waterbirds for all three encoders, whereas the curves cross on NICO++.}
\label{fig:gapcurve}
\end{figure}
Write $S=s(X)$ and assume finitely many groups with $\Pr(G=g)>0$. Pooled coverage is the mixture $C(\tau)=\sum_g\Pr(G=g)C_g(\tau)$.
\begin{proposition}[Equal coverage across thresholds]
\label{prop:independence}
$\Delta(\tau)=0$ for every threshold $\tau$ if and only if $S$ is independent of $G$.
\end{proposition}
At any threshold, an average equals its minimum only if all group coverages agree. Requiring this at every threshold is equivalent to equality of the conditional score distributions. The statement does not exclude parity at particular thresholds for a group-dependent score. It also does not equate group coverage with group accuracy. This is the independence criterion of the fairness literature \citep{barocas2023fairmlbook}, stated for abstention with its converse.

\subsection{Group leakage}
\label{sec:leakage}
Let us define $A_g$ to be the AUROC of $s$ for distinguishing group $g$ from its complement, which is the probability that a randomly selected member of g has a higher score than a randomly selected member of the complement of $g$, with ties counting as one half \citep{mann1947test,hanley1982meaning}. We distinguish the two-sided leakage, defined as, $\max_g |A_g-\tfrac12|$, from one-sided leakage defined as, $\max_g(\tfrac12-A_g)$. The latter captures groups that tend to score lower than the rest, which are therefore less likely to be accepted at a common threshold.
We denote these summaries by $L_{\pm}$ and $L_{\downarrow}$, respectively.
\begin{lemma}[cf.\ \citealp{traub2024evaluation}]
\label{lem:area}
On a fixed sample, sweep the threshold and parameterize it by pooled coverage $c\in[0,1]$. Let $\tau_c$ denote the corresponding threshold, $C_g(\tau_c)$ the coverage of group $g$, and $\pi_g$ its sample share. With fractional inclusion at tied scores:
\begin{equation}
\int_0^1 \bigl(c-C_g(\tau_c)\bigr)\,dc
= (1-\pi_g)\bigl(\tfrac12-A_g\bigr).
\label{eq:area}
\end{equation}
\end{lemma}
Thus, the group-versus-rest AUROC determines the signed area of the group's coverage deficit over the sweep of thresholds. The area is positive if the group is ranked below the rest on average, though its deficit can vary in sign at individual coverages. At a particular coverage, $\Delta(\tau_c)=\max_g\{c-C_g(\tau_c)\}$ is the maximum such deficit. Therefore:
\begin{equation}
\int_0^1 \Delta(\tau_c)\,dc
\geq \max_g (1-\pi_g)\bigl(\tfrac12-A_g\bigr).
\end{equation}
The above identity concerns the average over coverage; $A_g$  alone does not determine the gap at any particular operating point.
Figure~\ref{fig:gapcurve} shows these curves for the margin and the maximum logit. Proofs are in Appendix~\ref{app:proofs}.

\subsection{Randomized margin baseline}
\label{sec:noise}
Adding random noise gradually eliminates the association between a score and group membership. It thus has the potential to narrow the coverage gap, but only at the cost of reducing the informativeness of acceptance about correctness. We take this trade-off as our baseline. Let $F$ be the empirical CDF of the margin $\delta$ on the calibration split and let $U\sim\mathrm{Uniform}(0,1)$. For $\lambda\in\{0,0.1,\ldots,1\}$, define
\begin{equation}
s_\lambda = (1-\lambda)F(\delta) + \lambda U.
\label{eq:noise}
\end{equation}
We threshold $s_\lambda$ using the same procedure as for all other scores and average its coverage gap and pooled risk over 20 draws of $U$.

\begin{definition}
A score is \emph{noise-dominated} at coverage $c$ if there exists a noised margin $s_\lambda$ with no larger coverage gap and no larger pooled risk, with at least one inequality strict.
\end{definition}

Algorithm~\ref{alg:audit} summarizes this audit, which we apply to every encoder, prompt protocol, and confidence score.

\begin{algorithm}[t]
\caption{Audit a frozen VLM under group-blind abstention}
\label{alg:audit}
\begin{algorithmic}[1]
\Require Frozen encoder and prompts; unlabeled calibration inputs $\mathcal C$; test triples $\mathcal T=\{(x_i,y_i,g_i)\}$; fixed scores $\mathcal S$ and coverages $\mathcal Q$
\State Compute logits, fixed predictions $\hat y_i$, and scores on $\mathcal C$ and $\mathcal T$
\For{each $s\in\mathcal S$ and $c\in\mathcal Q$}
  \State $\tau_{s,c}\gets$ the $\lceil c|\mathcal C|\rceil$-th largest calibration score
  \State $a_i\gets\mathbf 1[s(x_i)\geq\tau_{s,c}]$ \Comment{same rule for every group}
  \State Evaluate $C,R,\{C_g,R_g\}_g,\Delta$ from $(a_i,\hat y_i,y_i,g_i)$
  \State Suppress worst-group risk if any group has fewer than 25 accepted inputs
  \State Obtain paired intervals against the margin by resampling within test groups
\EndFor
\State Compute group-versus-rest AUROCs and threshold-sweep diagnostics
\State Fit the margin CDF on $\mathcal C$; evaluate 11 noise mixtures, with 20 draws each, at $c=.8$
\State Compare each score with the noise mixtures jointly on gap and pooled risk
\State On Waterbirds, compare paired background swaps with same-type controls
\end{algorithmic}
\end{algorithm}

\section{Results and Analysis}
\label{sec:results}
We report test outcomes at calibration target $c=.8$, using Waterbirds A as the primary protocol and fixed predictions within each encoder--protocol pair. Appendix~\ref{app:appendix} reports protocol B, $c\in\{.7,.9\}$, and achieved test coverage, which can differ from its target.

\subsection{The cost of reducing the coverage gap varies by dataset}

On Waterbirds A, maximum logit reduces coverage gaps from 13.3, 22.0, and 15.6 percentage points (pp) to 4.2, 4.0, and 6.2 pp for OpenAI CLIP, LAION CLIP, and SigLIP, respectively. Margin gaps are 2.5--5.5 times as large; reduction intervals exclude zero in all nine encoder--coverage comparisons. At $c=.8$, pooled risk rises by 1.5--2.3 pp, while intervals for worst-group risk changes include zero (Table~\ref{tab:headline}). At $c=.9$, SigLIP's worst-group risk rises by 2.7 pp, with an interval excluding zero.

Relative to the group-agnostic reference in Section~\ref{sec:protocol}, the margin's worst-group risk has a Wilson interval above the reference in five of six Waterbirds encoder--protocol pairs. For OpenAI A, these risks are 28.1\% and 22.7\%. Maximum logit exhibits no such excess.

\begin{table}[!htbp]
\centering\small
\caption{Maximum logit minus margin at target coverage $c=.8$, in percentage points, with paired 95\% bootstrap intervals. Negative values indicate a reduction. Waterbirds uses species prompts (A). Intervals are pointwise and uncorrected for multiple comparisons.}
\label{tab:headline}
\setlength{\tabcolsep}{5pt}
\begin{tabular}{@{}llccc@{}}
\toprule
Dataset & Encoder & Coverage gap $\Delta$ & Worst-group risk & Pooled risk\\
\midrule
Waterbirds A & OpenAI CLIP & $-9.1$ {\scriptsize$[-12.2, -5.3]$} & $-2.0$ {\scriptsize$[-5.3, +1.3]$} & $+2.3$ {\scriptsize$[+1.6, +3.1]$} \\
 & LAION CLIP & $-17.9$ {\scriptsize$[-20.9, -13.7]$} & $-0.3$ {\scriptsize$[-4.0, +3.3]$} & $+1.9$ {\scriptsize$[+1.2, +2.6]$} \\
 & SigLIP & $-9.4$ {\scriptsize$[-12.4, -6.2]$} & $+1.8$ {\scriptsize$[-0.2, +4.0]$} & $+1.5$ {\scriptsize$[+1.1, +2.0]$} \\
\midrule
NICO++ & OpenAI CLIP & $+0.3$ {\scriptsize$[-1.0, +2.2]$} & $+5.4$ {\scriptsize$[+4.4, +6.5]$} & $+4.3$ {\scriptsize$[+3.9, +4.7]$} \\
 & LAION CLIP & $-2.6$ {\scriptsize$[-3.6, -0.6]$} & $+6.6$ {\scriptsize$[+5.4, +7.7]$} & $+4.1$ {\scriptsize$[+3.7, +4.6]$} \\
 & SigLIP & $-4.3$ {\scriptsize$[-5.6, -2.9]$} & $+6.6$ {\scriptsize$[+5.6, +7.5]$} & $+3.9$ {\scriptsize$[+3.6, +4.3]$} \\
\bottomrule
\end{tabular}
\end{table}

On NICO++, the margin's gap is 6.1--9.6 pp. Maximum logit changes it by $+0.3$, $-2.6$, and $-4.3$ pp, but increases worst-context risk by 5.4--6.6 pp and pooled risk by 3.9--4.3 pp. The gap curves cross (Figure~\ref{fig:gapcurve}), and the Waterbirds trade-off does not generalize. For neither score does the worst context's Wilson interval lie wholly above the group-agnostic reference. The margin also has lower AUGRC than maximum logit in all nine encoder--protocol pairs.

These effects remain when using an 80-template prompt ensemble, max or mean species pooling and ViT-L/14 encoders for both CLIP recipes. Maximum logit has the smaller Waterbirds gap in all 30 variants, with intervals excluding zero in 27, and the higher NICO++ worst-context risk in all 10 (Appendix~\ref{app:ablation}). On Waterbirds A with the larger OpenAI encoder, maximum logit raises worst-group risk by 7.4 pp $[4.0,10.7]$.

\subsection{Coverage gains must be compared with randomization}
\begin{figure}[!htbp]
\centering
\includegraphics[width=0.96\linewidth]{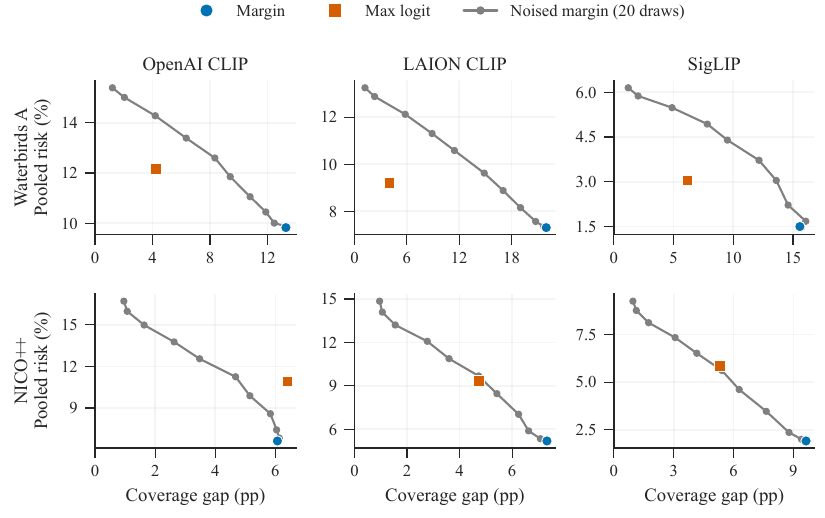}
\caption{Gap--risk trade-offs at $c=.8$. Gray curves connect means over 20 independent noise draws at each of 11 mixing weights. Lower and farther left is preferable; axes differ across panels.}
\label{fig:noise}
\end{figure}
Maximum logit is not noise-dominated in any Waterbirds encoder--protocol pair; under A, its risk is 2.1--3.3 pp below the interpolated noise envelope at the same gap. On NICO++, it is dominated for OpenAI. LAION and SigLIP lie within 0.4 pp of the interpolated envelope but are not dominated by an observed grid point (Figure~\ref{fig:noise}).

\subsection{The margin reflects object--background agreement}
\begin{figure}[!htbp]
\centering
\includegraphics[width=0.98\linewidth]{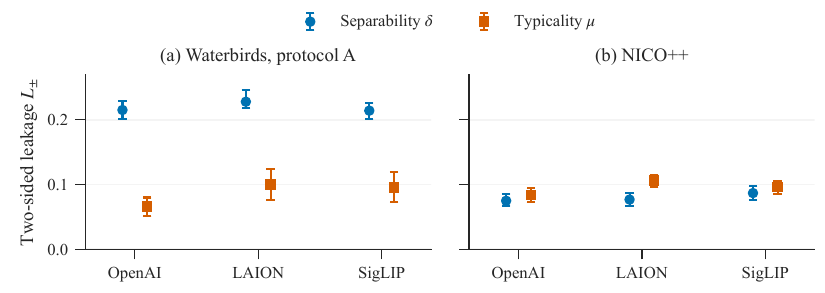}
\caption{Two-sided group leakage of separability and typicality, with stratified bootstrap 95\% intervals. Separability carries the larger group signal on Waterbirds; on NICO++ the ordering reverses weakly, and only LAION's paired difference resolves it.}
\label{fig:components}
\end{figure}
Across both Waterbirds protocols, separability's two-sided leakage is 0.21--0.31, versus 0.07--0.12 for typicality; all paired differences exclude zero. Under A, the margin distinguishes background-aligned from off-background images with AUROC 0.69--0.75, versus 0.58--0.61 for maximum logit. Its AUROC remains 0.67--0.72 among correct predictions, consistent with object--background agreement. NICO++ weakly reverses the leakage ordering, resolved only for LAION (Figure~\ref{fig:components}).

\subsection{Paired background changes suggest a qualified mechanism}
\label{sec:swap}
\begin{figure}[!htbp]
\centering
\includegraphics[width=\linewidth]{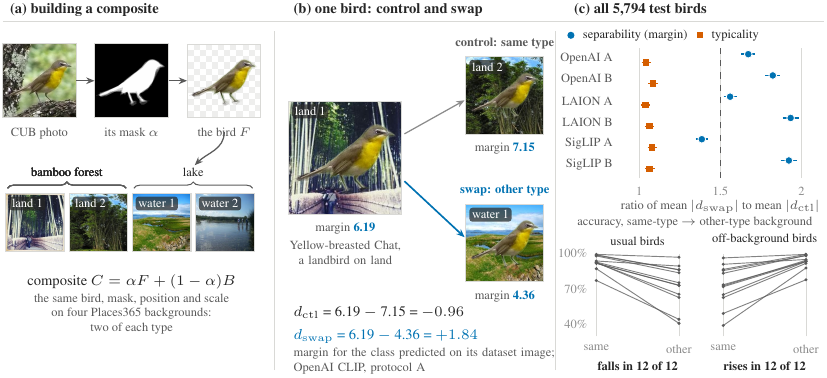}
\caption{Paired background swaps. Margins are signed for the original predicted class. The bird in (b) has both margin changes near the median for landbirds on land. In (c), intervals are 95\% over source birds; the dashed line marks the 1.5 criterion.}
\label{fig:swapdiagram}
\end{figure}
Across 23,176 composites with fixed foreground, position, and scale, we compare background-type swaps $d_{\rm swap}$ with same-type controls $d_{\rm ctl}$ (Figure~\ref{fig:swapdiagram}; Appendix~\ref{app:swap}). Class-aligned swaps improve off-background accuracy in all 12 encoder--protocol--group comparisons, and moving usual-background birds to the other type reduces their accuracy in all 12.

Here, the margin is signed relative to the original predicted class, so it can differ from the top-two margin after a flip. The ratio $\mathbb E|d_{\rm swap}|/\mathbb E|d_{\rm ctl}|$ is 1.67 $[1.63,1.72]$, 1.56 $[1.52,1.60]$, and 1.39 $[1.35,1.43]$ for OpenAI A, LAION A, and SigLIP A, respectively. All comparisons except SigLIP A meet the lower-interval criterion of 1.5. Fixed-class margin ratios (1.39--1.93) exceed typicality ratios (1.04--1.09), pointing to the margin, not typicality, as the main channel for background type. Scene variation and compositing artifacts limit causal interpretation.
\subsection{Direction matters, and the association survives duplicate removal}
\begin{table}[!htbp]
\centering\small
\caption{Spearman association with coverage gap at $c=.8$. Distinct entries omit the rank-normalized margin and identical separability score; the last row also omits coarse agreement scores. Shared models and data make cells dependent.}
\label{tab:correlation}
\begin{adjustbox}{max width=\linewidth}
\begin{tabular}{@{}lrrr@{\qquad}rrr@{}}
\toprule
& \multicolumn{3}{c}{Waterbirds (A and B)} & \multicolumn{3}{c}{NICO++}\\
Score set & $n$ & $L_{\downarrow}$ & $L_{\pm}$ & $n$ & $L_{\downarrow}$ & $L_{\pm}$\\
\midrule
All score entries & 108 & 0.880 & 0.611 & 54 & 0.919 & 0.649 \\
Distinct score entries & 96 & 0.873 & 0.670 & 48 & 0.921 & 0.676 \\
Without coarse scores & 84 & 0.926 & 0.724 & 42 & 0.959 & 0.738 \\
\bottomrule
\end{tabular}
\end{adjustbox}
\end{table}
The stronger downward association survives duplicate removal (Table~\ref{tab:correlation}) and holds within all nine encoder--protocol pairs, consistent with Lemma~\ref{lem:area}. Appendix~\ref{app:appendix} reports additional scores and the score-family sweep.

\section{Conclusion}
\label{sec:conclusion}
Confidence score choice changes how evenly groups are covered under a single group-blind threshold. On Waterbirds, separability carries more background-related group information than typicality, and maximum logit reduces coverage gaps while increasing overall selective risk. On NICO++, maximum logit increases worst-context risk and is noise-dominated for OpenAI CLIP. Downward group ordering tracks under-coverage more closely than two-sided group identifiability. These findings support evaluating group coverage and selective risk jointly and comparing apparent coverage gains against randomized baselines. Our evidence is limited to the tested models and contexts (Appendix~\ref{app:limitations}); whether these diagnostics predict score performance in new settings remains an open question.

\subsection*{Ethics statement}
Our experiments use publicly available benchmarks in accordance with their terms of use. These are Waterbirds (drawn from CUB-200-2011 and Places) and NICO++ (which allows research and non-commercial use). We generate background-swapped images from CUB and Places exclusively for analytic purposes. We have not collected new data from people.
The groups in this study refer to backgrounds and photographic contexts, not demographic characteristics,
and group labels are used only to evaluate abstention, never to determine it. Since coverage patterns differ between datasets, these results should not be interpreted as evidence of equitable abstention in settings involving people. Such applications require direct audits of group coverage.

\subsection*{Reproducibility statement}
Section~\ref{sec:setup} describes the encoders, prompts, calibration splits, threshold selection, and coverage grid. Section~\ref{sec:freeze} records the frozen protocol and the bootstrap. Scores and audit are included in Section~\ref{sec:scores} and Algorithm~\ref{alg:audit}, and the full definitions of scores appear in Appendix~\ref{app:scores}. The proofs of Proposition~\ref{prop:independence} and Lemma~\ref{lem:area} appear in Appendix~\ref{app:proofs}. No model is trained. We load public OpenCLIP checkpoints by name; Appendix~\ref{app:ablation} lists all five. Random draws are made
using a single recorded seed, which is stored with a hash of the full configuration. We verify the
Waterbirds release using a SHA-256 fingerprint of all 11,789 original files and we verify NICO++
image counts for each context and split. Pipeline reruns matched earlier results with numerical
precision across all shared tables, including a rerun by a different author on a different machine.

\FloatBarrier

\appendix
\section{Appendix}
\label{app:appendix}
\subsection{Limitations}
\label{app:limitations}
The study covers two datasets, three main checkpoints (two larger ones in Appendix~\ref{app:ablation}) and object/context groups. It does not measure training-seed variability or demographic fairness. Noise comparisons use a finite grid, post-test leakage analyses are descriptive, and synthetic background changes may differ from natural interventions.
\subsection{Proofs}
\label{app:proofs}
Assume each group has probability $\pi_g=\Pr[G=g]>0$ and $\sum_g\pi_g=1$.

\begin{proof}[Proof of Proposition~\ref{prop:independence}]
Conditioning on $G$ gives the pooled coverage:
\begin{equation}
C(\tau)=\Pr[S\ge\tau]=\sum_g\pi_g\Pr[S\ge\tau\mid G=g]
=\sum_g\pi_g C_g(\tau).
\label{eq:pf-mean}
\end{equation}
Define $m(\tau)=\min_g C_g(\tau)$. Since the shares add to one:
\begin{equation}
\Delta(\tau)=C(\tau)-m(\tau)
=\sum_g\pi_g\bigl(C_g(\tau)-m(\tau)\bigr).
\label{eq:pf-gapsum}
\end{equation}
All the terms are non-negative, and each share is positive, so:
\begin{equation}
\Delta(\tau)=0
\iff C_g(\tau)=C(\tau)\quad\text{for all }g.
\label{eq:pf-equal}
\end{equation}
If the gaps vanish at every threshold, then we also have:
\begin{equation}
\Pr[S\ge\tau\mid G=g]=\Pr[S\ge\tau]
\quad\text{for every }g\text{ and }\tau\in\mathbb{R}.
\label{eq:pf-survival}
\end{equation}
These equalities pin down the conditional distributions because for any $t$:
\[
\Pr[S\le t\mid G=g]
=1-\lim_{\tau\downarrow t}\Pr[S\ge\tau\mid G=g]
=\Pr[S\le t].
\]
Therefore, $S$ is independent of $G$. On the other hand, independence implies $C_g(\tau)=C(\tau)$ for all groups and thresholds, so that Equation~\ref{eq:pf-equal} shows $\Delta(\tau)=0$ everywhere.
\end{proof}

\begin{proof}[Proof of Lemma~\ref{lem:area}]
The argument follows the derivation of AUGRC in \citet[Eq.~8]{traub2024evaluation}, with group membership in place of classification failure. Fix a group $g$ with $0<\pi_g<1$ and set:
\begin{equation}
x(\tau)=C_{\neg g}(\tau)=\Pr[S\ge\tau\mid G\ne g],
\qquad y(\tau)=C_g(\tau).
\label{eq:pf-xy}
\end{equation}
As the threshold decreases, $(x,y)$ moves from $(0,0)$ to $(1,1)$. We interpolate jumps from tied scores linearly, which assigns a point on this path to every pooled coverage level. Since $x$ and $y$ are the acceptance rates outside and inside $g$, the path is the ROC curve for identifying membership in $g$ from $S$. Its area, with ties given half weight, is \citep{mann1947test,hanley1982meaning}: 
\begin{equation}
\int y\,dx=A_g.
\label{eq:pf-auc}
\end{equation}
The endpoints give, by integration by parts,
\begin{equation}
\int x\,dy=[xy]_{(0,0)}^{(1,1)}-\int y\,dx=1-A_g.
\label{eq:pf-parts}
\end{equation}
Also, $\int x\,dx=\int y\,dy=\tfrac12$. Therefore:
\begin{align}
\int(x-y)\,dx&=\tfrac12-A_g,\label{eq:pf-dx}\\
\int(x-y)\,dy&=\tfrac12-A_g.\label{eq:pf-dy}
\end{align}
Pooled coverage and the deficit for group $g$ satisfy:
\begin{equation}
c=\pi_g y+(1-\pi_g)x,
\label{eq:pf-c}
\end{equation}
and hence:
\begin{equation}
c-y=(1-\pi_g)(x-y),
\qquad dc=\pi_g\,dy+(1-\pi_g)\,dx.
\label{eq:pf-dc}
\end{equation}
Using $y=C_g(\tau_c)$ and integrating along the ROC path now yields:
\begin{align}
\int_0^1\bigl(c-C_g(\tau_c)\bigr)\,dc
&=(1-\pi_g)\int(x-y)\bigl[\pi_g\,dy+(1-\pi_g)\,dx\bigr] \nonumber\\
&=(1-\pi_g)\bigl[\pi_g(\tfrac12-A_g)
 +(1-\pi_g)(\tfrac12-A_g)\bigr] \nonumber\\
&=(1-\pi_g)(\tfrac12-A_g).\nonumber
\end{align}
\end{proof}

\begin{proof}[Proof of the bound in Section~\ref{sec:leakage}]
For any group $g$, its coverage cannot be less than the minimum group coverage. Hence for each pooled coverage level $c$:
\begin{equation}
\Delta(\tau_c)=c-\min_h C_h(\tau_c)\ge c-C_g(\tau_c).
\label{eq:pf-pointwise}
\end{equation}
Integrating this inequality and applying Lemma~\ref{lem:area} yields:
\begin{equation}
\int_0^1\Delta(\tau_c)\,dc
\ge\int_0^1\bigl(c-C_g(\tau_c)\bigr)\,dc
=(1-\pi_g)\bigl(\tfrac12-A_g\bigr).
\end{equation}
This holds for every $g$, and the left side does not depend on $g$. Therefore:
\begin{equation}
\int_0^1\Delta(\tau_c)\,dc
\ge\max_g(1-\pi_g)\bigl(\tfrac12-A_g\bigr).
\end{equation}
\end{proof}
\subsection{Score definitions}
\label{app:scores}

All scores use the base prediction $\hat y=\arg\max_k\ell_k$ from Section~\ref{sec:models}. For any class-logit vector $\ell'$, define its fixed-class margin by $m(\ell')=\ell'_{\hat y}-\max_{k\ne\hat y}\ell'_k$. Thus $m(\ell)=\ell_{(1)}-\ell_{(2)}$. Under protocol A, species logits are pooled into two class logits by logsumexp prior to any score being computed, for template, image-view and scene-prompt variants. Additional logits come from 80 CLIP templates $\ell^{(t)}$ \citep{radford2021clip}, four image views $\ell^{(v)}$ or scene clauses $z$ in a 16-clause vocabulary $Z$, yielding $\ell^{(z)}$ from ``a photo of a [class], $z$.'' The views are a horizontal flip, centre crops keeping 90\% and 80\% of each side and a flipped 90\% crop. In Table~\ref{tab:scores}, $F$ is the empirical CDF of its argument fit on the calibration split. All scores are oriented so higher values are accepted first. 

\begin{table}[!htbp]
\centering\footnotesize
\setlength{\tabcolsep}{4pt}
\renewcommand{\arraystretch}{1.1}
\caption{Scores and their information requirements beyond one image pass with the base prompts. The score family $s_\beta$ and noised margin $s_\lambda$ are defined in Section~\ref{sec:scores}.}
\label{tab:scores}
\begin{tabularx}{\linewidth}{@{}>{\raggedright\arraybackslash}p{0.21\linewidth}>{\raggedright\arraybackslash}X>{\raggedright\arraybackslash}p{0.22\linewidth}@{}}
\toprule
Score & Definition & Access \\
\midrule
Margin & $m(\ell)=\ell_{(1)}-\ell_{(2)}$ & output only \\
Maximum logit & $\ell_{(1)}$ & output only \\
Negative energy & $\log\sum_k \exp\ell_k$ \citep{liu2020energy} & output only \\
Typicality $\mu$, separability $\delta$ & $\tfrac12(\ell_{(1)}+\ell_{(2)})$ and $\ell_{(1)}-\ell_{(2)}$ (Equation~\ref{eq:components}). Separability equals the margin & output only \\
Rank-normalized margin & $F\bigl(m(\ell)\bigr)$, with the same ranking as the margin & output only \\
SAGE probe & $\max_t\bigl(\max_k\ell^{(t)}_k-\min_k\ell^{(t)}_k\bigr)$, the largest template-wise class separation, adapted from \citet{ye2026sage} & 80 prompts per class \\
Template variance & $-\operatorname{Var}_t m(\ell^{(t)})$ & 80 prompts per class \\
Template consistency & $\frac{1}{80}\sum_t\mathbf 1[\arg\max_k\ell^{(t)}_k=\hat y]$ \citep{ge2023improving}, with at most 81 values & 80 prompts per class \\
View agreement & $\frac14\sum_v\mathbf 1[\arg\max_k\ell^{(v)}_k=\hat y]$, with at most 5 values & 4 extra image passes \\
View mean & Mean of $m(\ell),m(\ell^{(1)}),\ldots,m(\ell^{(4)})$ \citep{bahat2018confidence} & 4 extra image passes \\
View variance & Negative variance of those five margins & 4 extra image passes \\
Context stability $s_{\rm CS}$ & $\tfrac12F\bigl(m(\ell)\bigr)+\tfrac12F(-I_T)$, where $I_T=Q_{0.9}-Q_{0.1}$ for $m(\ell^{(z)})$ over $z\in Z$ & 16 prompts per class \\
Rank-normalized stability & $F(-I_T)$ & 16 prompts per class \\
\bottomrule
\end{tabularx}
\end{table}

We evaluate context stability and its stability term using three vocabularies, giving 18 entries per encoder–protocol pair in Table~\ref{tab:correlation}. The deployed vocabulary was balanced on Waterbirds validation and then fixed for NICO++. Its clauses are \emph{on the ocean, on a lake, by a river, in a marsh, on a beach, in a harbor, on a pond, at the seashore, in a forest, in a bamboo grove, on a tree branch, in a meadow, in a garden, in the woods, on a mountain, in a field}. The initial vocabulary had only three water-related clauses and was rejected on validation. It used \emph{indoors, in a forest, on the water, on a street, in the snow, in a desert, in the mountains, on grass, in the sky, on a beach, in a city, in a field, on rocks, in a garden, in a harbor, in a meadow}. A control vocabulary has similarly formed clauses without scene meaning, including \emph{in a flarn} and \emph{near a quoxel}.

Template consistency and view agreement take only a small set of values. For two classes, maximum softmax probability and negative entropy have exactly the margin’s ranking and are therefore omitted. On NICO++, log maximum softmax probability $\ell_{(1)}-\log\sum_k\exp\ell_k$ has a Spearman correlation of 0.97--0.99 with the margin and is noise-dominated for OpenAI CLIP and SigLIP.

\subsection{Coverage and selective risk}
\begin{figure}[!htbp]
\centering
\includegraphics[width=\linewidth]{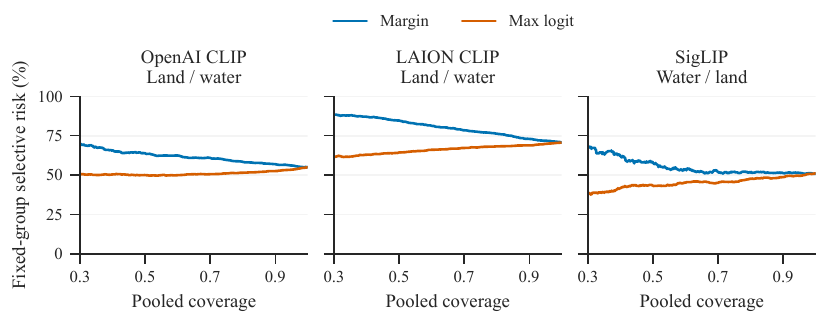}
\caption{Risk within each encoder's fixed, full-coverage worst group on Waterbirds B. Lowering coverage can increase the margin's accepted error rate. These are test-threshold sweeps; points with fewer than 25 accepted group images are omitted.}
\label{fig:group-risk}
\end{figure}

Table~\ref{tab:absolute} shows achieved coverage, coverage gap, worst-group risk and pooled risk for the margin and maximum logit for the three calibration targets. Figure~\ref{fig:group-risk} shows how, for each encoder, the full-coverage worst group on Waterbirds B progresses as the threshold is swept.
\begin{table}[!htbp]
\centering\small
\setlength{\tabcolsep}{3pt}
\caption{Absolute test results at three calibration targets. $C$ is achieved coverage, $\Delta$ coverage gap, $W$ worst-group risk, and $R$ pooled risk. Values are percentages, with gaps in percentage points. Achieved test coverage need not equal its calibration target.}
\label{tab:absolute}
\begin{adjustbox}{max width=\linewidth}
\begin{tabular}{@{}lllrrrrrrrr@{}}
\toprule
& & & \multicolumn{4}{c}{Margin} & \multicolumn{4}{c}{Maximum logit}\\
Dataset & Encoder & Target & $C$ & $\Delta$ & $W$ & $R$ & $C$ & $\Delta$ & $W$ & $R$\\
\midrule
Waterbirds A & OpenAI CLIP & 0.7 & 69.9 & 17.9 & 24.0 & 6.9 & 70.2 & 7.4 & 22.8 & 10.2 \\
Waterbirds A & OpenAI CLIP & 0.8 & 80.3 & 13.3 & 28.1 & 9.8 & 79.9 & 4.2 & 26.1 & 12.2 \\
Waterbirds A & OpenAI CLIP & 0.9 & 90.9 & 6.8 & 30.6 & 12.6 & 90.3 & 1.8 & 27.6 & 13.9 \\
\addlinespace[2pt]
Waterbirds A & LAION CLIP & 0.7 & 70.4 & 26.6 & 21.4 & 5.0 & 68.5 & 6.9 & 22.9 & 7.4 \\
Waterbirds A & LAION CLIP & 0.8 & 81.9 & 22.0 & 25.7 & 7.3 & 79.7 & 4.0 & 25.4 & 9.2 \\
Waterbirds A & LAION CLIP & 0.9 & 90.4 & 13.4 & 30.0 & 10.0 & 92.4 & 2.1 & 29.5 & 11.4 \\
\addlinespace[2pt]
Waterbirds A & SigLIP & 0.7 & 70.6 & 21.2 & 3.5 & 0.7 & 69.4 & 6.8 & 5.0 & 2.1 \\
Waterbirds A & SigLIP & 0.8 & 80.5 & 15.6 & 5.3 & 1.5 & 81.3 & 6.2 & 7.1 & 3.0 \\
Waterbirds A & SigLIP & 0.9 & 90.0 & 11.2 & 6.7 & 2.8 & 91.4 & 4.2 & 9.5 & 4.2 \\
\addlinespace[2pt]
Waterbirds B & OpenAI CLIP & 0.7 & 68.7 & 12.2 & 60.8 & 24.6 & 69.2 & 5.8 & 50.4 & 25.9 \\
Waterbirds B & OpenAI CLIP & 0.8 & 77.3 & 9.2 & 58.9 & 26.0 & 77.2 & 5.7 & 51.2 & 26.7 \\
Waterbirds B & OpenAI CLIP & 0.9 & 88.7 & 4.8 & 57.2 & 27.9 & 89.8 & 2.3 & 52.5 & 27.5 \\
\addlinespace[2pt]
Waterbirds B & LAION CLIP & 0.7 & 69.5 & 7.8 & 78.6 & 34.7 & 70.6 & 3.7 & 67.2 & 35.5 \\
Waterbirds B & LAION CLIP & 0.8 & 78.8 & 4.9 & 76.7 & 36.2 & 80.9 & 1.9 & 68.2 & 36.5 \\
Waterbirds B & LAION CLIP & 0.9 & 89.8 & 3.4 & 73.0 & 37.2 & 91.3 & 1.7 & 68.9 & 37.0 \\
\addlinespace[2pt]
Waterbirds B & SigLIP & 0.7 & 68.6 & 25.1 & 52.7 & 10.3 & 69.3 & 4.4 & 44.9 & 15.6 \\
Waterbirds B & SigLIP & 0.8 & 77.9 & 20.4 & 52.3 & 12.2 & 79.5 & 3.7 & 47.5 & 16.2 \\
Waterbirds B & SigLIP & 0.9 & 87.0 & 12.3 & 51.2 & 14.6 & 89.5 & 3.5 & 48.5 & 17.4 \\
\addlinespace[2pt]
NICO++ & OpenAI CLIP & 0.7 & 69.1 & 9.0 & 6.6 & 3.8 & 69.6 & 7.4 & 14.1 & 9.0 \\
NICO++ & OpenAI CLIP & 0.8 & 78.8 & 6.1 & 11.3 & 6.6 & 80.2 & 6.4 & 16.7 & 10.9 \\
NICO++ & OpenAI CLIP & 0.9 & 89.5 & 2.8 & 18.0 & 11.4 & 90.7 & 4.6 & 20.0 & 13.4 \\
\addlinespace[2pt]
NICO++ & LAION CLIP & 0.7 & 69.5 & 8.9 & 5.3 & 2.9 & 70.2 & 6.6 & 13.0 & 7.9 \\
NICO++ & LAION CLIP & 0.8 & 79.0 & 7.3 & 8.8 & 5.2 & 80.3 & 4.7 & 15.4 & 9.3 \\
NICO++ & LAION CLIP & 0.9 & 88.9 & 3.9 & 14.9 & 9.2 & 89.9 & 3.8 & 17.7 & 11.4 \\
\addlinespace[2pt]
NICO++ & SigLIP & 0.7 & 69.1 & 10.2 & 1.4 & 1.1 & 70.5 & 7.2 & 7.4 & 4.7 \\
NICO++ & SigLIP & 0.8 & 79.0 & 9.6 & 2.6 & 1.9 & 80.4 & 5.3 & 9.1 & 5.8 \\
NICO++ & SigLIP & 0.9 & 89.2 & 5.7 & 6.6 & 4.1 & 90.2 & 3.3 & 10.8 & 7.0 \\
\addlinespace[2pt]
\bottomrule
\end{tabular}
\end{adjustbox}
\end{table}

\subsection{Background changes and alternative scores}
\label{app:swap}
\begin{figure}[!htbp]
\centering
\includegraphics[width=\linewidth]{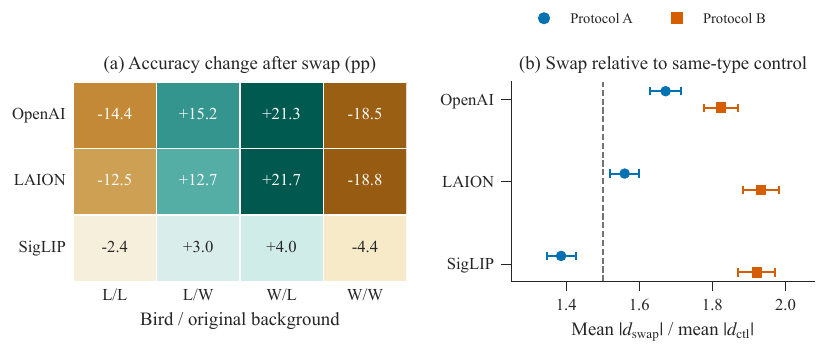}
\caption{Swaps use CUB foregrounds \citep{wah2011cub} and Places365 backgrounds \citep{zhou2017places}. Left: accuracy changes under protocol A; L/W denotes landbird on water. Right: mean absolute fixed-class margin change for a type swap divided by its same-type control, with source-bird bootstrap intervals. SigLIP A fails the lower-interval criterion of 1.5.}
\label{fig:swap}
\end{figure}

Figure~\ref{fig:swap} shows accuracy changes for each group under protocol A and the `swap to control' ratio for both protocols. Table~\ref{tab:baselines} contains four more scores at $c=.8$: negative energy, the SAGE probe, the view mean and context stability.

\begin{table}[!htbp]
\centering\footnotesize
\setlength{\tabcolsep}{2.5pt}
\caption{Alternative scores at calibration target $c=.8$. Each block reports gap $\Delta$, worst-group risk $W$, and pooled risk $R$ (\%, gaps in pp). All scores keep the base predictions. SAGE denotes a separation-based confidence probe; view mean uses four extra image passes. Context stability ($s_{\rm CS}$) gives no consistent gain.}
\label{tab:baselines}
\begin{adjustbox}{max width=\linewidth}
\begin{tabular}{@{}llrrrrrrrrrrrr@{}}
\toprule
& & \multicolumn{3}{c}{Energy} & \multicolumn{3}{c}{SAGE probe} & \multicolumn{3}{c}{View mean} & \multicolumn{3}{c}{$s_{\rm CS}$}\\
Dataset & Encoder & $\Delta$ & $W$ & $R$ & $\Delta$ & $W$ & $R$ & $\Delta$ & $W$ & $R$ & $\Delta$ & $W$ & $R$\\
\midrule
WB A & OpenAI & 3.5 & 26.0 & 12.5 & 15.6 & 32.4 & 9.0 & 13.5 & 22.8 & 7.6 & 19.8 & 31.0 & 9.9 \\
WB A & LAION & 4.0 & 25.4 & 9.2 & 20.2 & 25.9 & 6.3 & 23.6 & 21.9 & 5.6 & 11.0 & 32.2 & 8.1 \\
WB A & SigLIP & 5.2 & 7.6 & 3.3 & 15.4 & 4.6 & 1.2 & 18.2 & 4.0 & 0.9 & 25.1 & 13.2 & 2.4 \\
\addlinespace[2pt]
WB B & OpenAI & 4.4 & 50.9 & 26.9 & 8.3 & 59.6 & 26.2 & 10.1 & 54.5 & 23.7 & 12.9 & 53.0 & 24.3 \\
WB B & LAION & 1.7 & 68.0 & 36.4 & 11.7 & 82.4 & 40.2 & 5.9 & 72.9 & 33.6 & 7.6 & 70.7 & 33.5 \\
WB B & SigLIP & 3.3 & 47.4 & 17.0 & 14.3 & 44.6 & 16.9 & 23.0 & 54.4 & 10.7 & 21.6 & 62.1 & 12.8 \\
\addlinespace[2pt]
NICO++ & OpenAI & 7.0 & 18.3 & 12.2 & 8.1 & 15.9 & 9.8 & 6.7 & 9.8 & 5.7 & 16.5 & 12.9 & 10.8 \\
NICO++ & LAION & 4.7 & 15.8 & 9.8 & 5.7 & 13.8 & 8.2 & 7.2 & 7.8 & 4.4 & 15.2 & 10.4 & 8.7 \\
NICO++ & SigLIP & 4.9 & 10.2 & 6.4 & 5.4 & 9.3 & 5.3 & 10.0 & 2.0 & 1.6 & 15.1 & 5.5 & 4.6 \\
\bottomrule
\end{tabular}
\end{adjustbox}
\end{table}

\begin{table}[!htbp]
\centering\footnotesize
\setlength{\tabcolsep}{4pt}
\caption{Group sizes. Waterbirds groups cross bird class and background; NICO++ groups are contexts. Calibration labels are not used by the threshold order statistic.}
\label{tab:groupcounts}
\begin{adjustbox}{max width=\linewidth}
\begin{tabular}{@{}lrr@{\qquad}lrr@{}}
\toprule
Waterbirds group & Calibration & Test & NICO++ context & Calibration & Test\\
\midrule
Landbird / land & 467 & 2,255 & Autumn & 635 & 1,819\\
Landbird / water & 466 & 2,255 & Dim & 821 & 2,498\\
Waterbird / land & 133 & 642 & Grass & 1,375 & 4,066\\
Waterbird / water & 133 & 642 & Outdoor & 1,142 & 3,338\\
& & & Rock & 722 & 2,213\\
& & & Water & 1,305 & 3,841\\
\midrule
Total & 1,199 & 5,794 & Total & 6,000 & 17,775\\
\bottomrule
\end{tabular}
\end{adjustbox}
\end{table}

\subsection{Encoders and ablations}
\label{app:ablation}

\begin{figure}[!htbp]
\centering
\includegraphics[width=\linewidth]{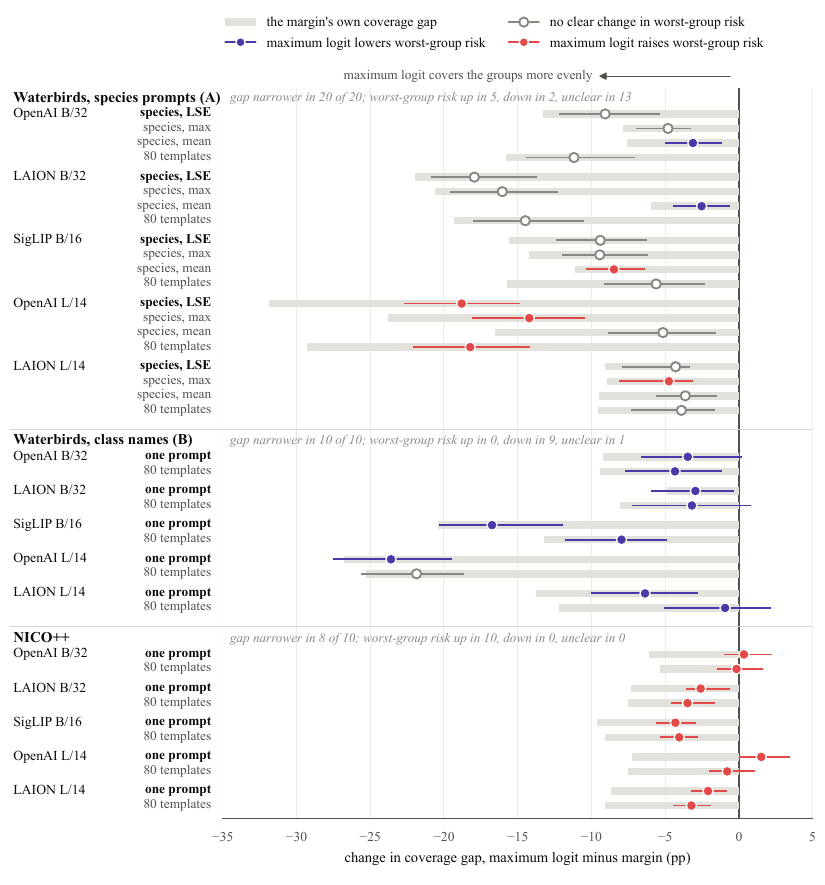}
\caption{Coverage gap of maximum logit minus that of the margin for every variant in Table~\ref{tab:ablation}, at $c=.8$, with paired 95\% bootstrap intervals. Bold rows are the classifiers of the main tables. The gray track runs from zero to minus the margin's own gap, so a dot at its left end would mean maximum logit leaves no gap. The fill shows how maximum logit changes worst-group risk: raised or lowered with an interval excluding zero, or no clear change (hollow). The Waterbirds gap narrows in all 30 variants, while on NICO++ it moves by at most 4.3 pp and worst-context risk rises in all 10.}
\label{fig:ablationforest}
\end{figure}

The primary experiments use three OpenCLIP checkpoints, \texttt{ViT-B-32-quickgelu}/\texttt{openai}, \texttt{ViT-B-32}/\texttt{laion2b\_s34b\_b79k} and \texttt{ViT-B-16-SigLIP}/\texttt{webli}. We additionally evaluate the larger \texttt{ViT-L-14-quickgelu}/\texttt{openai} and \texttt{ViT-L-14}/\texttt{laion2b\_s32b\_b82k} checkpoints under the respective OpenAI and LAION training recipes. Ablations of the classifier average each class or species prompt over 80 CLIP templates, and for protocol A also replace logsumexp species pooling with maximum or mean pooling. These variations of the classifier are therefore excluded from the main tables. Threshold selection, evaluation, bootstrap procedure and noise rule are the same. Table~\ref{tab:ablation} reports results at $c=.8$, and Figure~\ref{fig:ablationforest} shows the change in coverage gap for every variant.
Across the 30 Waterbirds variants, maximum logit always has a smaller coverage gap and higher pooled risk than the margin. The gap-difference interval excludes zero in 27 variants. Encoder size has no consistent effect on the margin’s gap in protocol A, which increases from 13.3 to 31.8 percentage points (pp) for OpenAI but decreases from 22.0 to 9.1 pp for LAION. For OpenAI L/14 under A, maximum logit increases worst-group risk by 7.4 pp $[4.0,10.7]$. It is noise-dominated in four Waterbirds variants, including three with LAION L/14. On NICO++ it increases worst-context risk in all 10 variants and is noise-dominated in every OpenAI variant. No test readout in the main analysis falls under the 25-image rule, or under a cutoff of 10 or 50. The fewest accepted images in any group are 134 on Waterbirds and 1,227 on NICO++.

In the background-swap experiment, the mean absolute change relative to the same-type control is 1.04--1.09 for typicality and 1.39--1.93 for separability. After scaling each component by its spread across dataset images, separability still changes more in five of six encoder--protocol pairs; SigLIP A is the exception, with a small advantage for typicality of $-0.03$ $[-0.05,-0.01]$.

\begingroup
\footnotesize
\setlength{\tabcolsep}{2.5pt}
\renewcommand{\arraystretch}{1.0}
\setlength{\LTleft}{0pt}
\setlength{\LTright}{0pt}
\setlength{\LTpre}{6pt}
\setlength{\LTpost}{6pt}
\begin{longtable}{@{}>{\raggedright\arraybackslash}m{0.13\linewidth}
>{\raggedright\arraybackslash}m{0.115\linewidth}
>{\raggedright\arraybackslash}m{0.12\linewidth}
>{\centering\arraybackslash}m{0.055\linewidth}
>{\centering\arraybackslash}m{0.055\linewidth}
>{\centering\arraybackslash}m{0.13\linewidth}
>{\centering\arraybackslash}m{0.13\linewidth}
>{\centering\arraybackslash}m{0.13\linewidth}
>{\centering\arraybackslash}m{0.03\linewidth}@{}}
\multicolumn{9}{@{}p{\textwidth}@{}}{\small\textbf{Table~\thetable:} Results at calibration target $c=.8$. Acc. gives classifier accuracy in percent and $\Delta_m$ gives the margin's coverage gap in pp. Gap and risk differences are maximum logit minus margin in pp, with paired 95\% bootstrap intervals. ND indicates that maximum logit is noise-dominated. Species with LSE and one prompt are the main classifiers. The 80-template variant averages each prompt across the CLIP templates.}\label{tab:ablation}\\[5pt]
\toprule
Data & Encoder & Classifier & Acc. & $\Delta_m$ & Gap & \shortstack[c]{Worst-group\\risk} & \shortstack[c]{Pooled\\risk} & ND\\
\midrule
\endfirsthead
\multicolumn{9}{@{}p{\textwidth}@{}}{\small\textbf{Table~\thetable:} Results at calibration target $c=.8$ (continued).}\\[5pt]
\toprule
Data & Encoder & Classifier & Acc. & $\Delta_m$ & Gap & \shortstack[c]{Worst-group\\risk} & \shortstack[c]{Pooled\\risk} & ND\\
\midrule
\endhead
\midrule
\multicolumn{9}{r}{\scriptsize Continued on the next page}\\
\endfoot
\bottomrule
\endlastfoot
Waterbirds A & OpenAI B/32 & species, LSE & 84.5 & 13.3 & \paperci{-9.1}{[-12.2, -5.3]} & \paperci{-2.0}{[-5.3, +1.3]} & \paperci{+2.3}{[+1.6, +3.1]} & no \\
Waterbirds A & OpenAI B/32 & species, max & 82.5 & 7.9 & \paperci{-4.8}{[-7.0, -3.2]} & \paperci{+0.4}{[-2.6, +1.9]} & \paperci{+2.0}{[+1.2, +2.8]} & no \\
Waterbirds A & OpenAI B/32 & species, mean & 69.5 & 7.6 & \paperci{-3.1}{[-5.0, -1.2]} & \paperci{-5.6}{[-7.2, -4.1]} & \paperci{+2.2}{[+1.4, +3.0]} & no \\
Waterbirds A & OpenAI B/32 & 80 templates & 84.3 & 15.8 & \paperci{-11.2}{[-14.5, -7.0]} & \paperci{-2.5}{[-5.9, +0.9]} & \paperci{+1.5}{[+0.7, +2.2]} & no \\
Waterbirds A & LAION B/32 & species, LSE & 86.7 & 22.0 & \paperci{-17.9}{[-20.9, -13.7]} & \paperci{-0.3}{[-4.0, +3.3]} & \paperci{+1.9}{[+1.2, +2.6]} & no \\
Waterbirds A & LAION B/32 & species, max & 86.4 & 20.6 & \paperci{-16.0}{[-19.6, -12.2]} & \paperci{-1.6}{[-5.2, +2.1]} & \paperci{+1.5}{[+0.8, +2.2]} & no \\
Waterbirds A & LAION B/32 & species, mean & 70.8 & 5.9 & \paperci{-2.5}{[-4.5, -0.6]} & \paperci{-2.2}{[-3.7, -0.8]} & \paperci{+2.2}{[+1.4, +3.0]} & no \\
Waterbirds A & LAION B/32 & 80 templates & 86.8 & 19.3 & \paperci{-14.5}{[-18.0, -10.5]} & \paperci{+0.5}{[-3.1, +4.3]} & \paperci{+1.6}{[+0.9, +2.4]} & no \\
Waterbirds A & SigLIP B/16 & species, LSE & 93.8 & 15.6 & \paperci{-9.4}{[-12.4, -6.2]} & \paperci{+1.8}{[-0.2, +4.0]} & \paperci{+1.5}{[+1.1, +2.0]} & no \\
Waterbirds A & SigLIP B/16 & species, max & 93.6 & 14.2 & \paperci{-9.4}{[-12.0, -6.1]} & \paperci{+1.0}{[-0.5, +3.0]} & \paperci{+1.5}{[+1.1, +2.0]} & no \\
Waterbirds A & SigLIP B/16 & species, mean & 80.1 & 11.1 & \paperci{-8.5}{[-10.4, -6.4]} & \paperci{+4.1}{[+2.5, +5.6]} & \paperci{+3.7}{[+2.9, +4.5]} & no \\
Waterbirds A & SigLIP B/16 & 80 templates & 94.2 & 15.7 & \paperci{-5.6}{[-9.2, -2.3]} & \paperci{+1.9}{[-0.2, +4.5]} & \paperci{+1.5}{[+1.0, +1.9]} & no \\
Waterbirds A & OpenAI L/14 & species, LSE & 95.0 & 31.8 & \paperci{-18.8}{[-22.7, -14.8]} & \paperci{+7.4}{[+4.0, +10.7]} & \paperci{+2.4}{[+1.9, +2.8]} & no \\
Waterbirds A & OpenAI L/14 & species, max & 94.6 & 23.8 & \paperci{-14.2}{[-18.1, -10.4]} & \paperci{+4.8}{[+1.7, +8.0]} & \paperci{+2.4}{[+1.9, +2.9]} & no \\
Waterbirds A & OpenAI L/14 & species, mean & 89.4 & 16.5 & \paperci{-5.2}{[-8.9, -1.6]} & \paperci{+1.9}{[-0.1, +5.0]} & \paperci{+3.3}{[+2.6, +4.0]} & no \\
Waterbirds A & OpenAI L/14 & 80 templates & 95.1 & 29.3 & \paperci{-18.2}{[-22.1, -14.2]} & \paperci{+6.6}{[+3.3, +10.0]} & \paperci{+2.1}{[+1.6, +2.5]} & no \\
Waterbirds A & LAION L/14 & species, LSE & 93.4 & 9.1 & \paperci{-4.3}{[-7.9, -3.3]} & \paperci{+1.5}{[-0.6, +3.8]} & \paperci{+1.5}{[+1.0, +2.0]} & no \\
Waterbirds A & LAION L/14 & species, max & 93.0 & 9.0 & \paperci{-4.7}{[-8.1, -3.1]} & \paperci{+2.2}{[+0.2, +4.4]} & \paperci{+1.8}{[+1.3, +2.3]} & no \\
Waterbirds A & LAION L/14 & species, mean & 76.7 & 9.5 & \paperci{-3.6}{[-5.7, -1.5]} & \paperci{-0.4}{[-2.0, +1.1]} & \paperci{+3.3}{[+2.5, +4.0]} & yes \\
Waterbirds A & LAION L/14 & 80 templates & 93.5 & 9.6 & \paperci{-3.9}{[-7.3, -1.6]} & \paperci{+0.2}{[-1.9, +2.4]} & \paperci{+1.3}{[+0.8, +1.8]} & no \\
\midrule
Waterbirds B & OpenAI B/32 & one prompt & 70.6 & 9.2 & \paperci{-3.5}{[-6.6, +0.2]} & \paperci{-7.6}{[-9.3, -6.0]} & \paperci{+0.7}{[-0.2, +1.5]} & no \\
Waterbirds B & OpenAI B/32 & 80 templates & 72.1 & 9.4 & \paperci{-4.3}{[-7.7, -1.1]} & \paperci{-5.8}{[-7.4, -4.1]} & \paperci{+1.1}{[+0.3, +2.0]} & no \\
Waterbirds B & LAION B/32 & one prompt & 61.6 & 4.9 & \paperci{-2.9}{[-6.0, -0.3]} & \paperci{-8.5}{[-10.0, -7.1]} & \paperci{+0.3}{[-0.6, +1.1]} & no \\
Waterbirds B & LAION B/32 & 80 templates & 62.1 & 8.0 & \paperci{-3.2}{[-7.3, +0.8]} & \paperci{-7.7}{[-9.0, -6.3]} & \paperci{+0.9}{[+0.0, +1.8]} & no \\
Waterbirds B & SigLIP B/16 & one prompt & 81.4 & 20.4 & \paperci{-16.7}{[-20.3, -11.9]} & \paperci{-4.8}{[-8.6, -1.1]} & \paperci{+4.0}{[+3.1, +4.8]} & no \\
Waterbirds B & SigLIP B/16 & 80 templates & 78.0 & 13.2 & \paperci{-8.0}{[-11.8, -4.9]} & \paperci{-5.4}{[-8.5, -2.2]} & \paperci{+4.6}{[+3.7, +5.4]} & yes \\
Waterbirds B & OpenAI L/14 & one prompt & 83.9 & 26.8 & \paperci{-23.6}{[-27.5, -19.5]} & \paperci{-5.8}{[-9.1, -2.5]} & \paperci{+4.8}{[+4.0, +5.5]} & no \\
Waterbirds B & OpenAI L/14 & 80 templates & 83.8 & 25.3 & \paperci{-21.9}{[-25.6, -18.7]} & \paperci{-3.3}{[-6.6, +0.2]} & \paperci{+4.9}{[+4.1, +5.6]} & no \\
Waterbirds B & LAION L/14 & one prompt & 74.2 & 13.8 & \paperci{-6.4}{[-10.0, -2.8]} & \paperci{-4.4}{[-6.0, -2.7]} & \paperci{+3.1}{[+2.3, +4.0]} & yes \\
Waterbirds B & LAION L/14 & 80 templates & 73.5 & 12.2 & \paperci{-0.9}{[-5.1, +2.2]} & \paperci{-3.8}{[-5.4, -2.1]} & \paperci{+3.7}{[+2.9, +4.6]} & yes \\
\midrule
NICO++ & OpenAI B/32 & one prompt & 83.3 & 6.1 & \paperci{+0.3}{[-1.0, +2.2]} & \paperci{+5.4}{[+4.4, +6.5]} & \paperci{+4.3}{[+3.9, +4.7]} & yes \\
NICO++ & OpenAI B/32 & 80 templates & 83.4 & 5.3 & \paperci{-0.2}{[-1.5, +1.6]} & \paperci{+6.1}{[+5.0, +7.2]} & \paperci{+4.2}{[+3.8, +4.7]} & yes \\
NICO++ & LAION B/32 & one prompt & 85.2 & 7.3 & \paperci{-2.6}{[-3.6, -0.6]} & \paperci{+6.6}{[+5.4, +7.7]} & \paperci{+4.1}{[+3.7, +4.6]} & no \\
NICO++ & LAION B/32 & 80 templates & 85.0 & 7.5 & \paperci{-3.5}{[-4.6, -1.6]} & \paperci{+7.9}{[+6.7, +9.0]} & \paperci{+4.8}{[+4.4, +5.2]} & no \\
NICO++ & SigLIP B/16 & one prompt & 90.8 & 9.6 & \paperci{-4.3}{[-5.6, -2.9]} & \paperci{+6.6}{[+5.6, +7.5]} & \paperci{+3.9}{[+3.6, +4.3]} & no \\
NICO++ & SigLIP B/16 & 80 templates & 90.9 & 9.1 & \paperci{-4.0}{[-5.4, -2.8]} & \paperci{+6.8}{[+5.8, +7.7]} & \paperci{+3.7}{[+3.4, +4.1]} & no \\
NICO++ & OpenAI L/14 & one prompt & 90.4 & 7.2 & \paperci{+1.5}{[-0.0, +3.5]} & \paperci{+4.0}{[+3.2, +4.8]} & \paperci{+3.1}{[+2.8, +3.5]} & yes \\
NICO++ & OpenAI L/14 & 80 templates & 90.8 & 7.5 & \paperci{-0.8}{[-2.0, +1.1]} & \paperci{+3.6}{[+2.9, +4.4]} & \paperci{+2.7}{[+2.4, +3.0]} & yes \\
NICO++ & LAION L/14 & one prompt & 90.6 & 8.6 & \paperci{-2.1}{[-3.3, -0.8]} & \paperci{+4.5}{[+3.7, +5.4]} & \paperci{+2.6}{[+2.3, +2.9]} & no \\
NICO++ & LAION L/14 & 80 templates & 91.1 & 9.0 & \paperci{-3.2}{[-4.5, -1.9]} & \paperci{+4.7}{[+3.8, +5.6]} & \paperci{+2.6}{[+2.3, +2.9]} & no \\
\end{longtable}
\endgroup
\subsection{Score-family and leakage diagnostics}
\begin{figure}[!htbp]
\centering
\includegraphics[width=.88\linewidth]{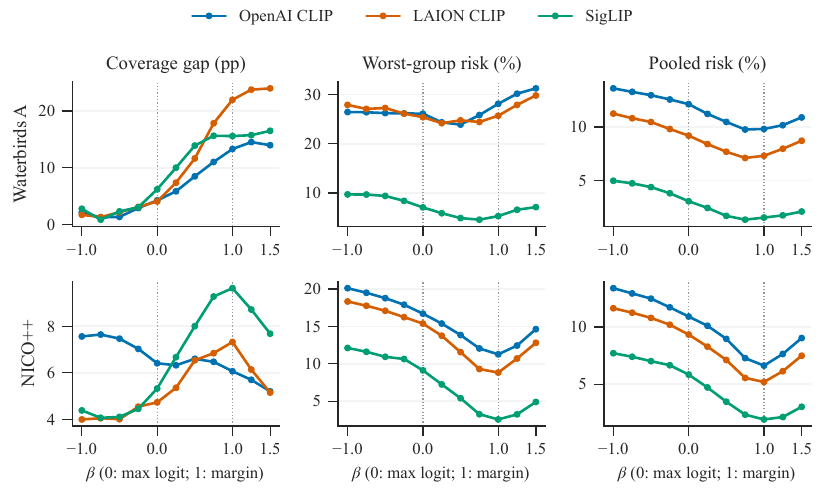}
\caption{Fixed score-family sweep at calibration target $c=.8$, with a separate threshold per score. Guides mark maximum logit ($\beta=0$) and margin ($\beta=1$). Gap and error need not share a minimizer; no test-optimal $\beta$ is selected.}
\label{fig:beta}
\end{figure}

\begin{figure}[!htbp]
\centering
\includegraphics[width=.88\linewidth]{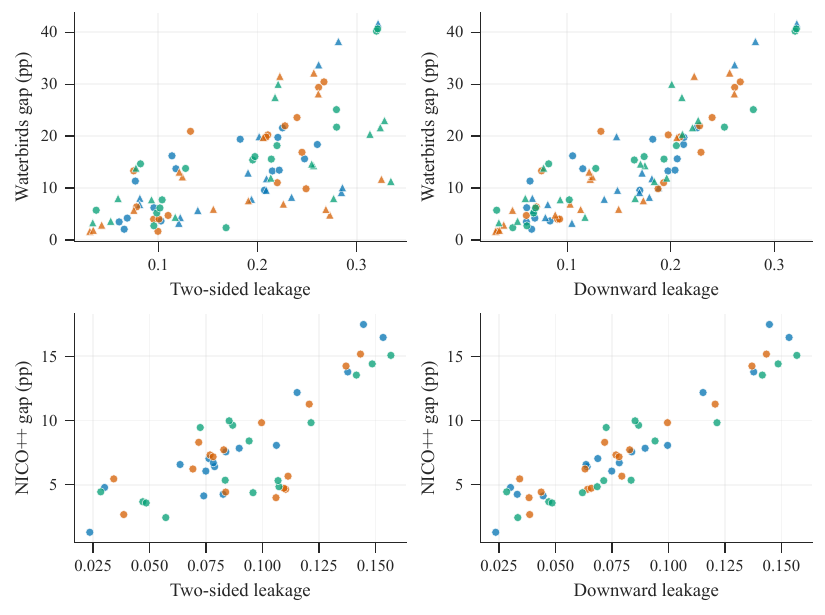}
\caption{Leakage versus coverage gap after removing margin copies. Blue, orange, and green denote OpenAI, LAION, and SigLIP. Circles denote Waterbirds A or NICO++, and triangles Waterbirds B. Downward leakage is more strongly associated with under-coverage; points share models and images and are not independent replications.}
\label{fig:leakage-scatter}
\end{figure}

Figure~\ref{fig:beta} depicts the score family $s_\beta$ of Equation~\ref{eq:family} with $c=.8$. Figure~\ref{fig:leakage-scatter} plots both leakage summarizations of Section~\ref{sec:leakage} against the coverage gap for the distinct score entries of Table~\ref{tab:correlation}. 

\end{document}